%% file: root.tex
\documentclass[letterpaper, 10 pt, conference]{ieeeconf}  

\IEEEoverridecommandlockouts                              

\usepackage{silence}
\usepackage[table,dvipsnames]{xcolor}      
\usepackage[noadjust]{cite}

\usepackage{amsmath,amssymb,amsfonts,amsthm,dsfont,mathtools}
\usepackage{multirow}
\usepackage{subcaption} 
\usepackage{svg}
\usepackage{graphicx}
\usepackage{graphicx,tabularx,adjustbox}
\graphicspath{{./fig/}}
\usepackage{tikz}
\usepackage[breaklinks=true, colorlinks, bookmarks=true]{hyperref}

\theoremstyle{definition}

\newtheorem*{problem*}{Problem}
\newtheorem{problem}{Problem}
\newtheorem*{remark*}{Remark}

\input{sym.tex}

\title{\LARGE \bf LATMOS: Latent Automaton Task Model from Observation Sequences}

\author{Weixiao Zhan$^1$ \and Qiyue Dong$^2$ \and Eduardo Sebasti\'an$^3$ \and Nikolay Atanasov$^2$
\thanks{We gratefully acknowledge support from ARL DCIST CRA W911NF-17-2-0181 and a US-Spain Fulbright grant.}%
\thanks{$^1$W. Zhan is with the Department of Computer Science and Engineering, University of California San Diego, La Jolla, CA 92093, USA (e-mail: \texttt{\small wezhan@ucsd.edu}).}%
\thanks{$^2$Q. Dong and N. Atanasov are with the Department of Electrical and Computer Engineering, University of California San Diego, La Jolla, CA 92093, USA (e-mail: \texttt{\small \{q4dong, natanasov\}@ucsd.edu}).}%
\thanks{$^3$E. Sebastián is with the Department of Computer Science and Technology, University of Cambridge, Cambridge, CB3 0FD, UK (e-mail: \texttt{\small es2121@cam.ac.uk}).}}

\newcommand\copyrighttext{%
  \footnotesize \textcopyright This paper has been accepted for presentation and publication at IEEE/RSJ  International Conference on Intelligent Robots and Systems (IROS), 2025. Please, when citing the paper, refer to the official manuscript.}
\newcommand\copyrightnotice{%
\begin{tikzpicture}[remember picture,overlay]
\node[anchor=south,yshift=10pt] at (current page.south) {\fbox{\parbox{\dimexpr\textwidth-\fboxsep-\fboxrule\relax}{\copyrighttext}}};
\end{tikzpicture}%
}

\begin{document}

\maketitle
\copyrightnotice

\begin{abstract}
Robot task planning from high-level instructions is an important step towards deploying fully autonomous robot systems in the service sector. Three key aspects of robot task planning present challenges yet to be resolved simultaneously, namely, 
(i) factorization of complex tasks specifications into simpler executable subtasks, 
(ii) understanding of the current task state from raw observations,  
and (iii) planning and verification of task executions. 
To address these challenges, we propose \textsf{LATMOS}, an automata-theory-inspired task model that, 
given observations from correct task executions, is able to factorize the task, while supporting verification and planning operations.
\textsf{LATMOS} combines an observation encoder to extract features from potentially high-dimensional observations with a sequence model that encapsulates an automaton with symbols in the latent feature space.
We conduct evaluations in three task model learning setups: (i) abstract tasks described by logical formulas, (ii) real-world human tasks described by videos and natural language prompts and (iii) a robot task described by image and state observations. The results show improved plan generation and verification capabilities of \textsf{LATMOS} across different observation modalities and tasks.
\end{abstract}

\input{Introduction.tex}

\input{ProblemStatement.tex}

\input{TechnicalApproach.tex}

\input{Evaluation.tex}

\input{Conclusion.tex}


\bibliographystyle{IEEEtran.bst}
\bibliography{IEEEabrv.bib,references.bib}

\end{document}

%% file: sym.tex
\DeclareMathOperator*{\argmax}{arg\,max}



%% file: Introduction.tex
\section{INTRODUCTION}
\label{sec:introduction}

Robot systems are increasingly integrated in unstructured human environments and expected to perform increasingly more complicated tasks autonomously. An important consideration is how to describe the objectives and constraints of a desired task to a robot, enabling human-understandable high-level specifications to be the main signal to command the robot systems. Task specification refers to the formal description of goals, constraints, and execution dependencies associated to a task, and it serves as a fundamental component of formulating autonomous robot behaviors, enabling robots to understand, decompose, and execute complex objectives while maintaining safety and failure recovery.

One popular formalism for expressing task specifications is Linear Temporal Logic (LTL) \cite{vardi2005automata, kress2007s, kress2009temporal, kantaros2022perception}. Every LTL formula is equivalent to a B\"uchi automaton \cite{mukund2012finite, smith2010optimal}, a class of automata that admits infinite input sequences and plays an important role in model checking \cite{Baier_PrinciplesModelChecking_Book08, kamale2021automata}. Automata-based task specification methods rely on explicit \textit{pre-defined} symbolic representation --- logical propositions associated to certain events for the system --- and automaton states and transitions, leading to interpretable graphical models that can be used either for task verification \cite{gu2024review} or planning \cite{araki2021learning, dai2024optimal}. However, constructing an automaton that represents a task requires substantial manual engineering effort and often restricts the symbols and task states to discrete spaces that lack the expressivity to capture real-world robotics scenarios involving high-dimensional image data and continuous robot states.

\begin{figure}[t]
    \centering
    \includegraphics[width=1\linewidth]{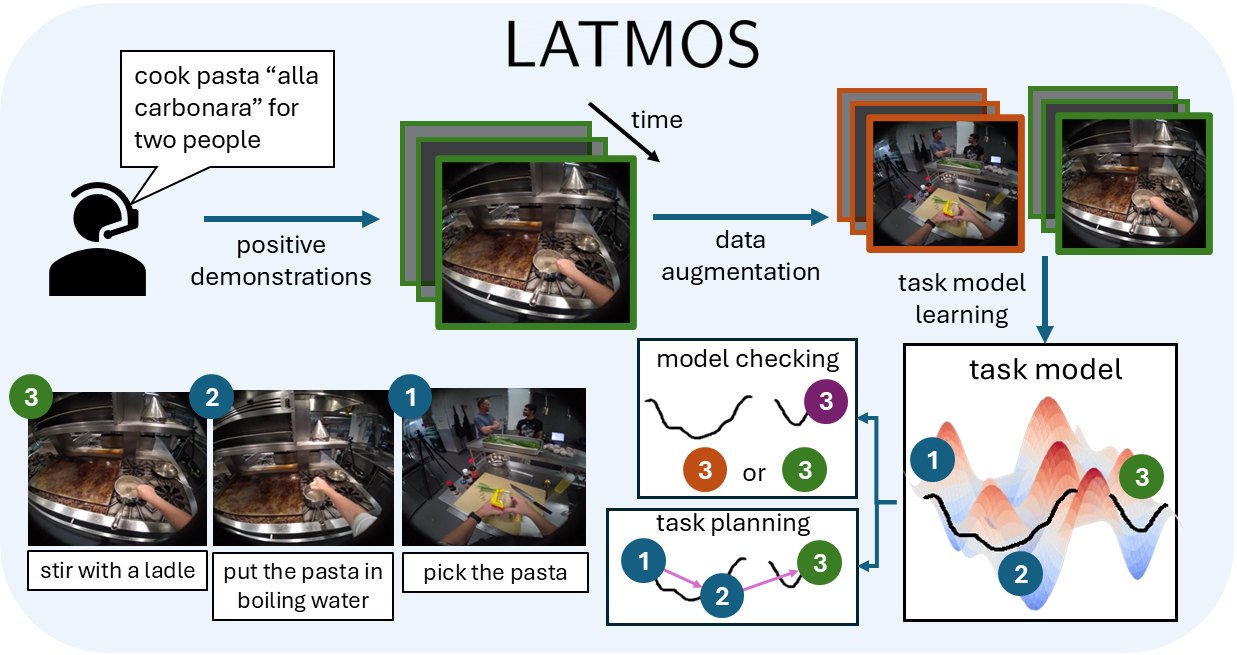}
    \caption{We present \textsf{LATMOS}, an automata-theory-inspired method to learn continuous task models from raw observations of positive task demonstrations. Given a dataset of task executions, our method first synthesizes negative demonstrations. Then, an encoder\(\rightarrow\)sequence-model\(\rightarrow\)decoder architecture learns a continuous latent-space representation of a task automaton. After training, the model is used for task plan verification and generation.}
    \label{fig:teaser}
\end{figure}


On the other hand, recent theoretical advances have established principled connections between Recurrent Neural Networks (RNNs) and Deterministic Finite Automata (DFA) \cite{Li_TheoreticalConnectionAutomataRNN_ML24}, suggesting the potential for learning continuous task representations. Concurrently, the success of pre-trained feature extractors \cite{girdhar2022omnivore, pramanick2023egovlpv2, ni2022expandinglanguageimagepretrainedmodels} offers the ability to capture complex patterns and relationships in continuous embedding spaces from images, videos and natural languages. However, there is still a gap to connect these theoretical results and practical insights to address robot task representation, in particular, in how to design a training signal from expert data to learn a meaningful automata-inspired task representation. Consequently, an important question is: \emph{given observations of task demonstrations, can we learn a continuous task representation that allows to validate a robot's execution of the task and plan ahead the next steps in the task space?}

We formalize this question in Sec.~\ref{sec:problem_statement} and propose an approach, termed as \textsf{LATMOS}, for learning latent task models from positive task demonstrations in Sec.~\ref{sec:technical_approach}. Our approach is illustrated in Fig. \ref{fig:teaser}. Instead of relying on handcrafted symbols and task states, \textsf{LATMOS} leverages multi-modal encoders to directly extract symbols from the observations. Then, a sequence model learns an internal state representation in a continuous domain that implicitly encodes the steps of the task. Lastly, a decoder model transforms the task state to a human interpretable signal. 
In Sec. \ref{sec:evaluation}, we demonstrate that \textsf{LATMOS} achieves superior performance in model checking and robot task planning compared to automata-based approaches across different environments and tasks using simulated and real data. These results show that \textsf{LATMOS} can be integrated effectively as a high-level decision-making component in the stack of autonomous robot systems. 


\section{Related Work}\label{sec:related}

Propositional logic \cite{meli2023logic}, temporal logic \cite{plaku2016motion}, and automata \cite{araki2019learning} provide a structured formulation of task specifications for robot systems. Since automata are graphical representations, they can be composed with the graph structures used in robot motion planning algorithms to achieve task planning \cite{vasile2020reactive,zhao2022reactive,li2023sampling,pan2024task}. Temporal logic specifications \cite{Kantaros_ReactiveMultirobotLTL_ICRA20, Sun_MultiAgentSTL_RAL22} capture the temporal ordering of events and can be converted to equivalent automata, enabling optimal control of long-horizon tasks in high-dimensional robot systems. Recent works incorporate semantic reasoning and natural language understanding into temporal logic task planners \cite{Fu_SemanticLTLPlanning_ICRA16, Ravichandran_SPINE_arXiv24, Chen_AutoTAMP_ICRA24, Guo_ConstraintsStL_arXiv24}, allowing robots to interpret high-level commands in unstructured environments. 
While temporal logic and automata models have a clear structure amenable to providing task execution guarantees, they rely on handcrafted specifications of propositions and task states. Our approach bypasses this by integrating learning with automata principles, allowing robots to \emph{learn} task structure, monitor execution states, and plan task transitions.

Traditional automata learning methods, like Angluin’s L* algorithm \cite{Angluin_Lstar_IC87} and its extensions \cite{Vazquez_LstarLLM_arXiv24}, are based on a tabular classification of sequences of symbols and require positive demonstrations and negative counterexamples from an oracle. Statistical approaches like ALERGIA \cite{Carrasco_ALERGIA_RAIRO99} overcome the need of counterexamples, using positive sequences of symbols to compute the transition probabilities among states of a stochastic finite automaton (SFA). Spectral methods \cite{Arrivault_SpLearn_CFAA17,Subakan_SpectralMHMM_NIPS14,Balle_SpectralWeightedAutomata_ICAI15} conduct a frequency analysis over the symbols to infer both the transition probabilities and the states of an SFA, enabling model verification and planning by selecting and comparing the transition probability given a state. Acknowledging the need for more flexible approaches, Li et al. \cite{Li_TheoreticalConnectionAutomataRNN_ML24} theoretically connect weighted automata, tensor networks, and recurrent neural networks via spectral learning, a connection that has not been fully exploited yet. Baert et al. \cite{baert2024learning} use an SFA to model task execution and, subsequently, learn the model from demonstrations. In the robot task planning domain, Araki et al. \cite{araki2019learning, araki2021learning} use automata as an internal representation of a ``logical'' Markov Decision Process that is called in a Value Iteration process for planning. Although these approaches exploit learning to add flexibility to automata-based task representations, they still rely on handcrafted symbols or other classes of logical templates. Instead, one of the focus of our work is to avoid manual engineering efforts and develop a method that automatically learns task models directly from observations. 

Recently, works on task planning based on Large Language Models (LLMs) demonstrate their ability to generate action sequences directly from observations and prompts without explicit task structures \cite{Brohan_RT2_arXiv2023, song2023llm, kannan2024smart}. While such approaches enable flexible planning, they often lack correctness guarantees due to their implicit reasoning. This limitation is mitigated by works that use world models \cite{guan2023leveraging} or hierarchical scene graphs \cite{ray2024task} to improve LLM-based planning. An alternative paradigm is to decompose task planning into explicit task, robot, and environment models, allowing for modularity, verifiable correctness and reliability. We propose \textsf{LATMOS} to learn a task model from observations that enables plan generation and verification in the task space rather than the action space, decoupling the task model from the robot model and action policy.

%% file: ProblemStatement.tex
\section{PROBLEM DEFINITION}
\label{sec:problem_statement}


This section introduces the definitions and data needed to formalize the problem of task model learning. Sec.~\ref{subsec:dfa} presents background on automata theory, Sec.~\ref{subsec:demonstrations} introduces the main features of the data assumed to be available, and Sec.~\ref{subsec:formulation} formalizes the problem. 

\subsection{Deterministic Finite Automata}\label{subsec:dfa}
A task specification can be encoded as a Deterministic Finite Automaton (DFA) operating in a discrete time \( k \in \mathbb{N} \). A DFA is a tuple \(\mathcal{A} = (S, \Sigma, \delta, s_0, F)\), where:
\begin{itemize}
    \item \(S\) is a finite set of states,
    
    \item \(\Sigma\) is a finite set of inputs, called alphabet,
    
    
    \item \(\delta: S \times \Sigma \to S\) is a transition function such that the next state is given by a transition \(s_{k+1} = \delta(s_k, o_k)\) dependent on the current state \(s_k \in S\) and input \(o_k \in \Sigma\),

    
    \item \(s_0 \in S\) is an initial state,
    
    \item \(F \subseteq S\) is a set of accepting (final) states, which model the completion of a task.
    
\end{itemize}
We assume that the automaton \(\mathcal{A}\) representing the task, including all of its elements \(S\), \(\Sigma\), \(\delta\), \(s_0\) and \(F\), is unknown.


\subsection{Positive Task Demonstrations}\label{subsec:demonstrations}
Instead of an automaton representation, we assume that the task of interest is described by observation sequences from successful task demonstrations. Specifically, let \(o^j_k \in \mathbb{R}^N\) be an \(N\)-dimensional observation from task demonstration \(j\) at time \(k\).
The sequence of observations obtained from one task execution is denoted by \mbox{\(o^j = \{o^j_k\}_{k=0}^{K_j}\)}, where \(K_j > 1\) is the time-horizon of the demonstration. 
%
The collection of observations from all demonstrations is denoted by \(O = \{o^j\}_{j=1}^L\), where \(L\) is the number of demonstrations.  

We emphasize that we assume the availability of positive demonstrations only. In general robotics applications, the space of unsatisfactory task executions is much larger than that of the positive demonstrations. Furthermore, negative demonstrations of a robotic task, such as assistance in a hospital setting, involves concerns and risks that cannot be taken. Therefore, in a safety-critical context, it is reasonable to assume the absence of negative samples in \(O\).

\subsection{Problem Definition}\label{subsec:formulation}

Our objective is to learn a task model with \emph{continuous} states given \emph{continuous} observations \(O\) from positive task demonstrations. The purpose of learning a continuous model instead of a discrete (automaton) model is to enable flexibility with respect to novel observations out of the distribution of the positive demonstrations contained in \(O\), and robustness against  perturbations that arise in unstructured environments typical of robotic applications. We also consider continuous states in the task model to allow learning a suitable latent state space and state transition function. Let \(\mathcal{M}_{\theta}(o^j)\) be a neural network model parameterized by \(\theta \in \mathbb{R}^M\) which takes an observation sequence \(o^j\) as input. We consider the following problem.

\begin{problem}\label{prob:problem_definition}
    Given positive task demonstrations \(O\), find parameters \(\theta^*\) such that the model \(\mathcal{M}_{\theta}(o^j)\) is a continuous representation of an automaton \(\mathcal{A}\) with the following capabilities.
\begin{enumerate}
    \item \textbf{Model checking}: Given an observation sequence \(o^j\), model \(\mathcal{M}_{\theta^*}\) can verify if \(o^j\) satisfies the task.
    
    \item \textbf{Task planning}: Given an observation sequence \(o^j\), model \(\mathcal{M}_{\theta^*}\) can predict an extension \(\tilde{o}^j\) such that the concatenated sequence \(o^j || \tilde{o}^j\) satisfies the task.

    
\end{enumerate}
\end{problem}

Having access to a task model with the properties in Problem~\ref{prob:problem_definition} would allow a robot system to plan a task execution and verify its correctness relying on sensor observations.


%% file: TechnicalApproach.tex
\section{Learning Task Representations for  Planning}
\label{sec:technical_approach}

\begin{figure}[t]
    \centering
    \includegraphics[width=1\linewidth]{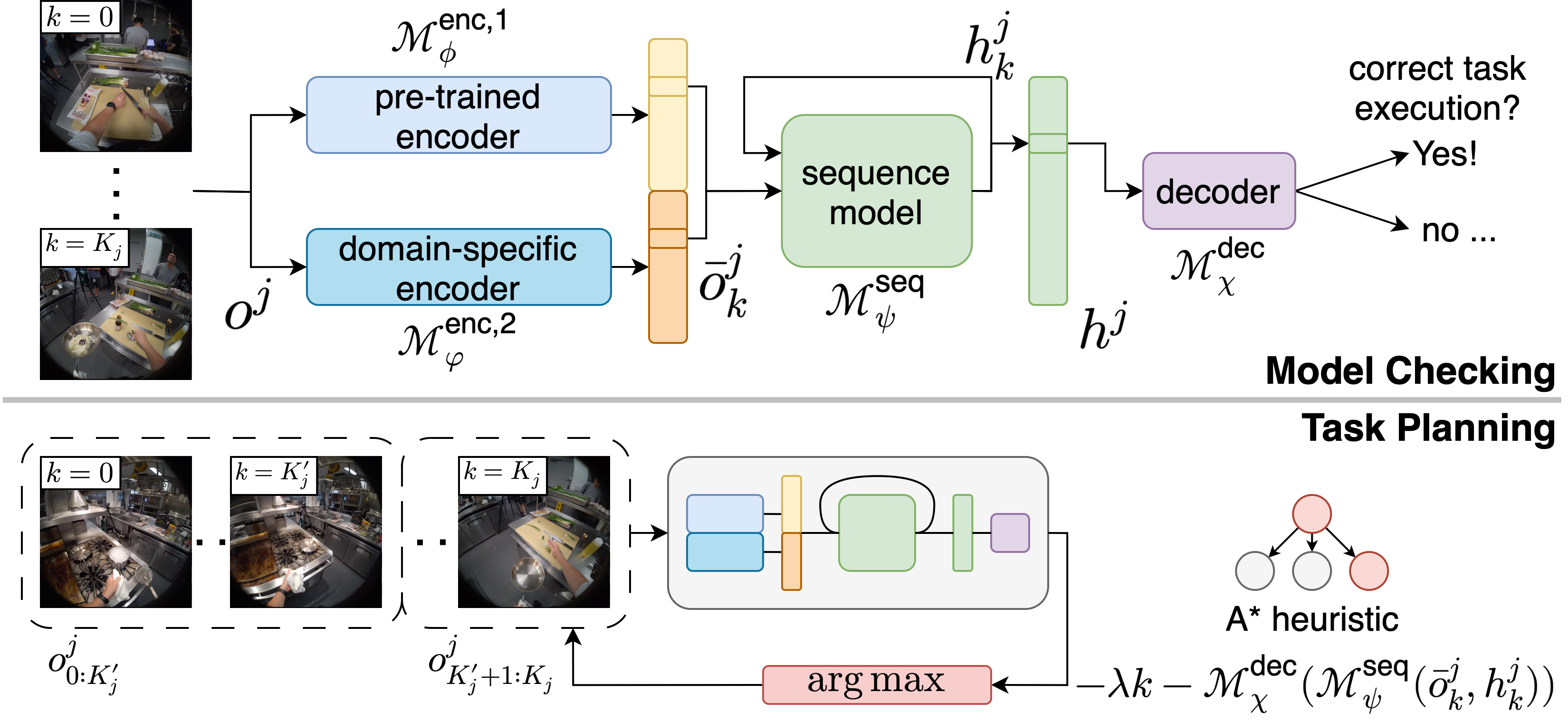}
    \caption{Overview of \textsf{LATMOS}.
    \textbf{(Top)} Given observation sequences from task demonstrations (e.g., cooking videos), a neural network encoder is used to extract features from each observation. The resulting feature sequences are used to train a sequence model and a decoder with a model-checking objective to solve the first item of Problem \ref{prob:problem_definition}. \textbf{(Bottom)} Once trained, the task model may be queried to obtain a heuristic function as guidance for a motion planning algorithm executed in the learned latent task space.}
    \label{fig:sys_overview}
\end{figure}

To solve Problem \ref{prob:problem_definition}, we propose \textsf{LATMOS} (\textsf{L}atent \textsf{A}utomaton \textsf{T}ask \textsf{M}odel from 
\textsf{O}bservation \textsf{S}equences), a method that combines the benefits of a neural network encoder to extract features from high-dimensional observations with a neural network sequence model to represent the task structure, and a neural network decoder to output the probability that an observation sequence comes from a correct task execution. Fig. \ref{fig:sys_overview} presents an overview of \textsf{LATMOS}. In Sec.~\ref{subsec:automata_learning}, we describe how to learn task models from just positive task demonstrations. In Sec.~\ref{subsec:vision}, we describe how to use a neural network encoder to overcome the need for handcrafted discrete input symbols as in an automaton representation. In Sec.~\ref{subsec:planning}, we derive a task planning method based on inference-time optimization, enabling task planning in the learned latent task space.

\subsection{Training Task Models Through Model Checking}
\label{subsec:automata_learning}

The first challenge of learning task representations solely from positive demonstrations is that the training data is inherently unbalanced. To tackle this challenge, we augment the observation dataset \(O\) with synthetically generated negative examples. For ease of exposition, we first assume that the set of observations \(O\) is formed by sequences from a known automaton alphabet \(\Sigma\). Later on, in Sec.~\ref{subsec:vision}, we describe how to deal with high-dimensional observations (e.g., images) and remove the assumption of a known alphabet.

For each sequence of observations \(o^j\), we select an observation \(o_k^j\), copy the prior observation sequence \(o^j_{0:k}\) and denote it by \(\tilde{o}^j_{0:k}\). 
Then, we sample a sequence of new synthetic observations \(\tilde{o}^j_{k+1:{K_j}}\) uniformly and independently from the alphabet \(\Sigma\). The sequences \(\tilde{o}^j_{0:k}\) and \(\tilde{o}^j_{k+1:{K_j}}\) are concatenated, leading to a synthetic negative demonstration \(\tilde{o}^j\). This step is repeated for each \(o^j_{k} \in o^j\). The synthetic negative demonstrations are included in a augmented dataset, denoted as \(\tilde{O}\), which also contains the original positive demonstrations in \(O\). After that, binary labels are assigned to the demonstrations. Label \(y^j_k = 1\) is assigned to all the observations \(o^j_k\) that belong to the original dataset \(O\) and those belonging to the prior sequences of observations \(o^j_{0:k}\) sampled to generate the negative synthetic sequence \(\tilde{o}^j\). Label \(y^j_k = 0\) is assigned to all the new synthetic observations \(\tilde{o}^j_{k+1:{K_j}}\).
This method produces a balanced dataset with observations from positive and negative task executions. Our synthetic label generation approach is feasible under the assumption that the space of negative demonstrations is exponentially larger than the space of positive demonstrations, such that most random observation sequences lead to an incorrect execution of the task. In practice, this assumption holds when the observation dimension \(N\) is very large, e.g., in the case of robot task executions with visual observations.

Given the augmented dataset \(\tilde{O}\), we pose the problem of learning \(\mathcal{M}_{\theta^*}\) as the minimization of the cross-entropy \(H(\hat{y}^j_k, y^j_k)\) between the prediction of the model \(\hat{y}^j_k\) and the ground-truth labels \(y^j_k\) associated to \(\tilde{O}\). 
We use one-hot encoding on the binary ground-truth labels \(y^j_k\). 
To obtain \(\hat{y}^j_k\), we parameterize \(\mathcal{M}_{\theta}\) by means of two neural networks. The first neural network is a sequence model \(\mathcal{M}^{\textsf{seq}}_\psi(o_k^j, h_k^j)\), parameterized by \(\psi\), that is in charge of learning a latent state representation of the automaton \(\mathcal{A}\), denoted as \(h^j_k\). The second neural network is a decoder \(\mathcal{M}^{\textsf{dec}}_\chi(h^j_{K_j})\), parameterized by \(\chi\), that takes the latent state \(h^j_k\) as input and outputs the probability of acceptance of the sequence of observation \(o^j_{0:k}\). The activation function of the last layer of the decoder is set to be SoftMax to obtain a vector \(\hat{y}^j_k\) containing the probability of acceptance and failure. 

Learning a continuous representation of the task from demonstrations then translates into finding parameters \(\psi\) and \(\chi\) that minimize the cross-entropy loss between ground-truth labels and predicted labels:
\begin{equation}\label{eq:contrastive}
\begin{aligned}
\min_{\psi,\chi}\quad & \mathbb{E}_{o^j_k\sim \tilde{O}} \left[ \frac{1}{L} \sum_{j=1}^L \frac{1}{K_j} \sum_{k=0}^{K_j} H\left( \hat{y}^j_k, y^j_k \right) \right]
\\
s.t. \quad & h^j_{k+1} = \mathcal{M}^{\textsf{seq}}_\psi(o_k^j, h_k^j) \quad \forall k,j,
\\
&\hat{y}^j_k = \mathcal{M}^{\textsf{dec}}_\chi(h^j_{k}) \quad \forall k,j.    
\end{aligned}
\end{equation}

In the next section, we extend the formulation to handle high-dimensional observations from an unknown \(\Sigma\).

\subsection{From Symbols to High-Dimensional Observations}\label{subsec:vision}

Task planning in real-world robotics applications cannot rely on hand-crafted finite alphabets because it is hard to choose a set of discrete symbols that describes all possible variations of high-dimensional observations, such as images from an onboard camera. We propose an encoder-based method to automatically infer relevant symbols from task demonstrations in the form of latent-space features.


Let \(o_k^j\) be a high-dimensional observation of the environment at time \(k\) from demonstration \(j\), e.g., an image or a set of frames grouped together in the same vector. Instead of directly using \(o_k^j\) as input to \(\mathcal{M}_{\psi}^{\mathsf{seq}}\), we use a neural network encoder to extract features from \(o_k^j\). Specifically, the encoder is a combination of a pre-trained model \(\mathcal{M}_{\phi}^{\mathsf{enc, 1}}(o_k^j)\) with parameters \(\phi\), and a small task-specific encoder \(\mathcal{M}_{\varphi}^{\mathsf{enc, 2}}(o_k^j)\) with parameters \(\varphi\). The outputs of the two models are concatenated to obtain \(\bar{o}_k^j\). The pre-trained encoder maps semantically similar inputs to nearby regions in the embedding space, making extracted features behave similarly to symbols of an automaton alphabet. Hence, \(\phi\) is frozen during training. On the other hand, the task-specific encoder is a smaller model that is jointly trained with the sequence model and decoder to adapt domain-specific observations.

Using the feature encoders, minimization \eqref{eq:contrastive} is reformulated as follows:
\begin{equation}\label{eq:complete}
\begin{aligned}
\min_{\psi,\chi, \varphi}\quad & \mathbb{E}_{o^j_k\sim \tilde{O}} \left[ \frac{1}{L} \sum_{j=1}^L \frac{1}{K_j} \sum_{k=0}^{K_j} H\left( \hat{y}^j_k, y^j_k \right) \right]
\\
s.t. \quad & \bar{o}^j_{k} =  \mathcal{M}^{\textsf{enc, 1}}_\phi(o_k^j) || \mathcal{M}^{\textsf{enc, 2}}_\varphi(o_k^j) \quad \forall k,j,
\\
& h^j_{k+1} = \mathcal{M}^{\textsf{seq}}_\psi(\bar{o}_k^j, h_k^j) \quad \forall k,j,
\\
&\hat{y}^j_k = \mathcal{M}^{\textsf{dec}}_\chi(h^j_{k}) \quad \forall k, j.    
\end{aligned}
\end{equation}
In \eqref{eq:complete}, the outputs of the encoder are interpreted as input symbols of an automatically inferred alphabet with \(N^{\prime}\) symbols. Unlike the symbols in a DFA, our model treats the embedding vectors as continuous symbols, which enables robustness against the variability of the real-world situations that robots encounter during deployment, e.g., different camera perspectives and object rearrangements.

In summary, the proposed model \(\mathcal{M}_{\theta}(o^j)\) to solve Problem \ref{prob:problem_definition} is composed of three modules: (i) an encoder (\(\mathcal{M}^{\textsf{enc, 1}}_\phi(o_k^j),  \mathcal{M}^{\textsf{enc, 2}}_\varphi(o_k^j)\)) that takes high-dimensional observations and learns to automatically extract the symbols of a relevant alphabet; (ii) a sequence model (\(\mathcal{M}^{\textsf{seq}}_\psi(\bar{o}_k^j, h_k^j)\)) that learns a latent state representation of the task; and (iii) a decoder (\(\mathcal{M}^{\textsf{dec}}_\chi(h^j_{K_j})\)) that learns to interpret the latent state of the sequence model and assess whether the task is solved. The trainable model parameters are \(\theta = \{\varphi, \psi, \chi\}\). By training the model \(\mathcal{M}_{\theta}(o^j)\) as in \eqref{eq:complete}, we address the model checking requirement in item 1 of Problem \ref{prob:problem_definition}.  We discuss how to use \(\mathcal{M}_{\theta}(o^j)\) for task planning to address item 2 of Problem \ref{prob:problem_definition} next.

\subsection{Latent Space Task Planning}
\label{subsec:planning}
 
\textsf{LATMOS} enables task planning through test-time optimization. Given an observation sequence \(o^j_{0:{K_j}^{\prime}}\) and our model \(\mathcal{M}_{\theta}(o^j)\), we pose task planning as the following optimization problem:
\begin{align}\label{eq:maximization}
    \argmax_{\bar{o}^j_{{K_j}^{\prime}+1:{K_j}^{\prime\prime}}} & \quad \mathcal{M}^{\textsf{dec}}_\chi(h^j_{{K_j}^{\prime\prime}})
    \\
    s.t. \quad & h^j_{k+1} = \mathcal{M}^{\textsf{seq}}_\psi(\bar{o}_k^j, h_k^j) \hbox{ } \forall k \in \{{K_j}^\prime, \hdots, {K_j}^{\prime\prime}\}, \notag
\end{align}
with \(\bar{o}^j_{{K_j}^{\prime}+1:{K_j}^{\prime\prime}} = \{\bar{o}_k^j\}_{k={K_j}^{\prime}}^{{K_j}^{\prime\prime}}\) and \({K_j}^{\prime\prime} > {K_j}^{\prime}\) the last planning step. To solve \eqref{eq:maximization}, we use a model-guided A* search \cite{hart1968formal}. To apply A*, we assume that we have access to a simulator or an observation model that can be queried to synthesize task paths (see, e.g., the 2D grid environment in Sec. \ref{subsec:doorkey}). Given an initial observation \(\bar{o}_{{K_j}^\prime}^j\) and associated latent hidden state \(h_{{K_j}^\prime}^j\), we define the following heuristic function:
\begin{equation}\label{eq:heuristic}
    \gamma(h_k^j,k) = -\lambda k - \mathcal{M}^{\textsf{dec}}_\chi(\mathcal{M}^{\textsf{seq}}_\psi(\bar{o}_k^j, h_k^j)),
\end{equation}
where \(\lambda\) trades off search exploration and current task path extension. The dynamic programming nature of A* allows efficient exploration of a space with exponentially many path by reusing previously computed latent hidden states. The output of the model-guided A* algorithm is a sequence of observations and associated latent states that solve the task. The observation sequence can then be used as input to a robot control algorithm to generate actions that lead to observations aligned with the desired observation sequence. 

Alternatively, since our task model is differentiable, it is possible to employ gradient-based optimization for task planning. As the exact time-horizon of the plan is not known a priori, one can set a pre-defined horizon \(T\) and do planning in the latent task space in a receding-horizon model predictive control fashion, leveraging \(\mathcal{M}_{\theta}(o^j)\) as a dynamic model of the task. We leave the analysis of this approach for future work. Both planning methods enable robots to exploit model \(\mathcal{M}_\theta(o^j)\) for task planning, even though it is trained using just a model checking loss from positive demonstrations. Hence, \textsf{LATMOS} addresses item 2 in Problem \ref{prob:problem_definition}.

%% file: Evaluation.tex
\section{EXPERIMENTS}
\label{sec:evaluation}

We evaluate \textsf{LATMOS} in three types of experiments\footnote{Code at: \url{https://github.com/weixiao-zhan/LATMOS}}:
\begin{enumerate}
    \item synthetic tasks specified by LTL formulas using Spot \cite{Michaud_Spot_WS18}, which allow to assess model checking with different sequence model configurations of \textsf{LATMOS} under known alphabets in Sec.~\ref{subsec:automata_learning};
    
    \item real-world tasks specified by video demonstrations using the Ego-Exo4D dataset \cite{Grauman_EgoExo4D_CVPR24}, which allow to asses the model checking properties of \textsf{LATMOS} when the alphabet is unknown and only raw high-dimensional observations are available in Sec.~\ref{subsec:vision};
    
    \item goal-conditioned tasks in a Door-Key grid environment \cite{chevalier2023minigrid}, which allows to assess the domain-specific encoder and task planning of \textsf{LATMOS} in Sec.~\ref{subsec:planning}.
\end{enumerate}
In all experiments, the decoder is parameterized as a Multi-Layer Perceptron (MLP) with one hidden layer whose dimension is the average of the input and output dimension. The MLP uses Leaky ReLU activations in the hidden layer. The decoder is deliberately simple to force the sequence model to learn a high-quality latent representation.

Since model checking with \textsf{LATMOS} reduces to binary classification of an input sequence \(O^j\), the evaluation metric for model checking is acceptance accuracy,
\begin{equation*}
\text{acceptance accuracy}=\frac{\text{TP}_j + \text{TN}_j}{\text{TP}_j + \text{TN}_j + \text{FP}_j + \text{FN}_j},   
\end{equation*}
where TP, TN, FP and FN refer to the number of true positives, true negatives, false positives and false negatives respectively. Regarding the task planning capabilities of \textsf{LATMOS}, we consider search efficiency as the evaluation metric:
\begin{equation*}
\text{search efficiency} = \frac{E_j}{{K_j}^{\prime\prime}},
\end{equation*}
where \(E_j\) is the number of explored states and \({K_j}^{\prime\prime}\) is the length of the shortest task plan achieved. For both metrics, we compute the average over the test set sequences.

\subsection{Automata Learning from Positive Demonstrations}  
\label{subsec:ltl_tasks}

\begin{figure}[t]
    \centering
    \includegraphics[width=\columnwidth]{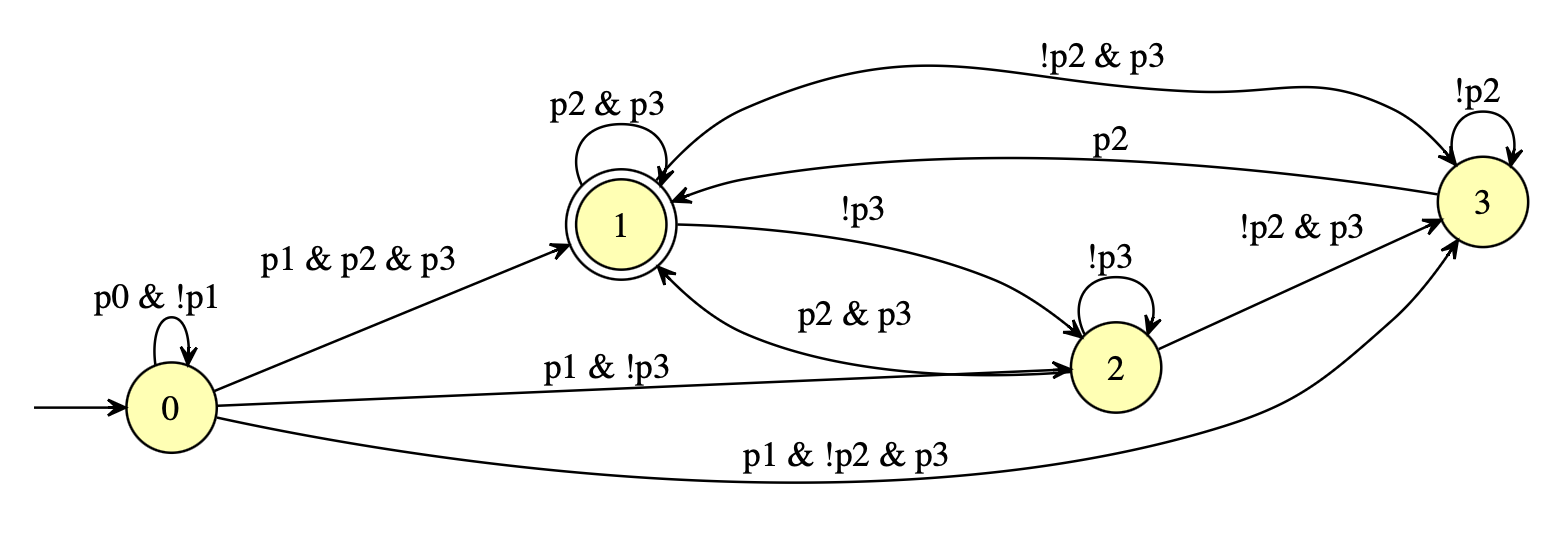}
    \caption{Example of ground-truth automaton with \(4\) APs and \(4\) states. State $1$ is the accepting state of the task.}
    \label{fig:example_automata}
\end{figure}

We first evaluate our method on synthetic tasks specified by randomly generated LTL formulas with corresponding ground-truth automata. To generate training and test datasets, we use \(3\) automata (see Fig. \ref{fig:example_automata}) and generate \(L=1000\) sequences from each via random walks on the automaton graphs, with varying sequence lengths up to \({K_j}=|S|\) to ensure enough exploration. Since we have access to the ground-truth automata, dataset augmentation is not needed and we directly employ the Atomic Propositions (APs) of the LTL formulas as observations. Three variations of the same LTL formula are tested:  
\begin{itemize}
    \item \textbf{Base}: Training and testing use exact APs collected by the random walks.  
    
    \item \textbf{Noisy APs}: Training APs are perturbed by additive zero-mean Gaussian noise with covariance \(\mathbf{Q} = \{0.1\mathbf{I}, 0.2\mathbf{I}\}\), where \(\mathbf{I}\) is the identity matrix of appropriate dimensions. Testing APs are kept exact.
    
    \item \textbf{Novel APs}: The test sequences contain transitions absent in the training dataset. 
\end{itemize}
We compare \textsf{LATMOS} against two traditional automata learning methods: \textsf{ALERGIA} \cite{mao2016ALERGIA} and Spectral Learning (\textsf{Sp-Learn}) \cite{Arrivault_SpLearn_CFAA17}. \textsf{ALERGIA} builds a look-up table during training and traverses the table to assess acceptance at test time. \textsf{Sp-Learn} constructs a Hankel matrix to capture the frequency of chains of APs and to learn the states \(S\) and the transition function \(\delta\). 
For \textsf{Sp-Learn}, the rank factor is the ratio between the number of states in the learned automaton, i.e., the rank of the Hankel matrix, and the actual number of states in the ground-truth automaton.
Regarding \textsf{LATMOS}, we ablate three most popular parametrizations of the sequence model \(\mathcal{M}_{\psi}^{\mathsf{seq}}\): a Gated Recurrent Unit (GRU) \cite{chung2014gru} network, an Attention-based network \cite{vaswani2017attention}, and a State-Space Model (SSM) \cite{gu2021ssm} network. All parametrizations are configured to have similar number of parameters and trained with the same number of epochs and learning rate. The \textit{hidden factor} is the ratio between the dimension of the hidden state and the number of states in the ground-truth automaton. 

Tables~\ref{tab:LTL_model_checking_0} and \ref{tab:LTL_model_checking_1}
show the acceptance accuracy for the different setups and model configurations. 
\textsf{Sp-Learn} performs the best when its Hankel Matrix has exactly the same rank as the number of states in ground-truth automata, which implies that it is a critical factor but it may be hard to tune in the absence of ground-truth automata. 
In contrast, \textsf{LATMOS} performs consistently with large-enough hidden dimensions. We note that an Attention network with a hidden factor of 0.5 may not be implemented due to model-size incompatibilities inherent to the multi-head attention architecture. On the base dataset (Table \ref{tab:LTL_model_checking_0}), all models achieve high accuracy, confirming their ability to learn automata. However, in the presence of noisy or novel APs (Table \ref{tab:LTL_model_checking_1}), \textsf{ALERGIA}, \textsf{Sp-Learn} completely fail due to their reliance on exact AP matching. On the other hand, \textsf{LATMOS} performs consistently in both cases. Specifically, the Attention backbone slightly outperforms GRU and SSM. This is deemed to the fact that Attention processes the whole observation sequence at each step; meanwhile, GRU and SSM are recurrent methods that process information step by step. These results validate that our learned latent representations capture automata structure while making meaningful interpolations and generalizing to noisy and unseen APs, which are critical properties for real-world robot applications where observations are imperfect.

\begin{table}[t]
\centering
\caption{Accuracy for the {base} configuration.}
\label{tab:LTL_model_checking_0}
\begin{tabular}{l|cccc}
\multirow{2}{*}{model configuration} & \multicolumn{4}{c}{rank/hidden factor} \\
\cline{2-5}
 & 0.5 & 1 & 4 & 12 \\
\hline
\textsf{ALERGIA} \cite{mao2016ALERGIA} & \multicolumn{4}{c}{\textbf{1.000}} \\
\textsf{Sp-Learn} \cite{Arrivault_SpLearn_CFAA17} & 0.834 & \textbf{0.948} & 0.945 & 0.925 \\
\textsf{LATMOS} (GRU) & 0.125 & 0.314 & 0.720 & \textbf{0.957} \\
\textsf{LATMOS} (Attention) & - & 0.148 & 0.956 & \textbf{1.000} \\
\textsf{LATMOS} (SSM) & 0.125 & 0.520 & 0.895 & \textbf{0.956} \\
\end{tabular}
\end{table}

\begin{table}[t]
\setlength{\tabcolsep}{4pt}
\centering
\caption{Accuracy for the {noisy} and {novel} configurations.}
\label{tab:LTL_model_checking_1}
\resizebox{\linewidth}{!}{\begin{tabular}{lcccc}
\multirow{2}{*}{model configuration} & \multirow{2}{*}{rank/hidden factor} & \multicolumn{3}{c}{accuracy}\\
\cline{3-5}
& & var=0.1 & var=0.2 & novel \\
\hline
\textsf{ALERGIA} \cite{mao2016ALERGIA} & - & 0 & 0 & 0\\
\textsf{Sp-Learn} \cite{Arrivault_SpLearn_CFAA17} & 1 & 0 & 0 & 0.639\\
\textsf{LATMOS} (GRU) & 12 & 0.775 & 0.668 & 0.686\\
\textsf{LATMOS} (Attention) & 12 & \textbf{0.794} & \textbf{0.681} & \textbf{0.864}\\
\textsf{LATMOS} (SSM) & 12 & 0.787 & 0.666 & 0.648
\end{tabular}}
\end{table}

\subsection{Automata Learning from Real-World Visual Cues}\label{subsec:egoexo}

Ego-Exo4D \cite{Grauman_EgoExo4D_CVPR24} is a diverse, large-scale multi-modal, multi-view video dataset that contains time-aligned egocentric and exocentric videos of people completing tasks like cooking or repairing a bike. Specifically, we use the key-step sub-dataset, in which the segments of the videos are annotated with step ID and natural-language captions. Sequences of segments constitute positive task demonstrations. In this experiment, three state-of-the-art pre-trained feature extractors are compared: XCLIP \cite{ni2022expandinglanguageimagepretrainedmodels}, Omnivore \cite{girdhar2022omnivore}, and EgoVLPv2 \cite{pramanick2023egovlpv2}.
The feature extractors are trained to handle Vision (V) or Language (L) modalities. XCLIP and Omnivore are trained in generalist datasets whereas EgoVLPv2 is specialized to Ego-Exo4D data. 
No task-specific encoder is used. For the data augmentation process, we leverage the encoder structure of \textsf{LATMOS} to directly sample in the embedding space. Following the insights gained from Sec.~\ref{subsec:ltl_tasks}, \(\mathcal{M}_{\psi}^{\mathsf{seq}}\) is set as an Attention-based model with \(4\) layers and \(4\) attention heads, with \(512\) hidden dimensions.

To further assess the benefits of \textsf{LATMOS}, we train a second key-step decoder to solve the single key-step recognition challenge included in Ego-Exo4D \cite{Grauman_EgoExo4D_CVPR24}. The goal of the challenge is to classify step IDs \(s_k^j\) from the visual observation segments \(o_k^j\). We adopted \textsf{LATMOS} for this challenge to study whether the learned latent states $h_k^j$ are representative of the task and help in improving prediction accuracy in the benchmark. The key-step decoder is parameterized as the original acceptance decoder except the output uses SoftMax activation with 629 dimension (the number of unique step IDs). We first train the sequence model with the acceptance decoder, as proposed in \eqref{eq:contrastive} and \eqref{eq:complete}; then, we train the key-step decoder with \(\mathcal{M}_{\psi}^{\mathsf{seq}}\) frozen. 

\begin{figure}[ht]
    \centering
    \begin{subfigure}[b]{0.45\columnwidth}
        \centering
        \includegraphics[width=\linewidth]{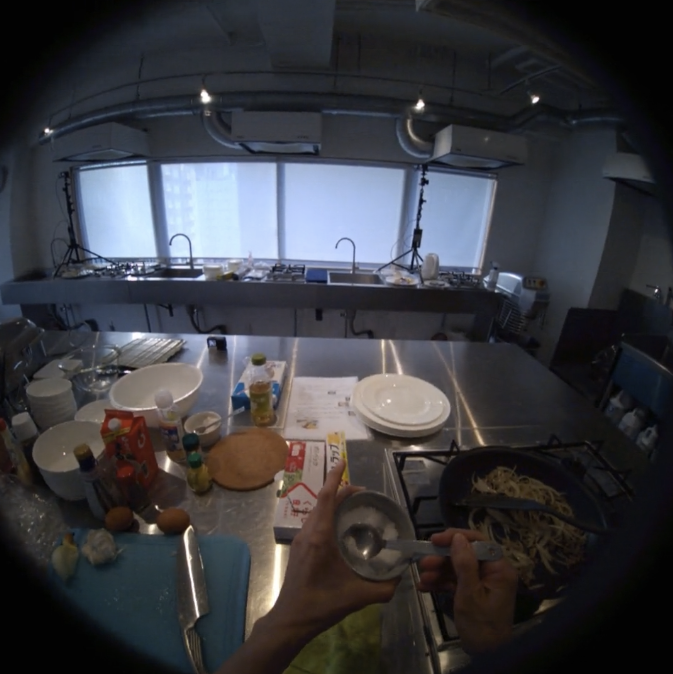}
        \caption{Caption: ``add salt''}
        \label{fig:fig1}
    \end{subfigure}
    \hfill
    \begin{subfigure}[b]{0.45\columnwidth}
        \centering
        \includegraphics[width=\linewidth]{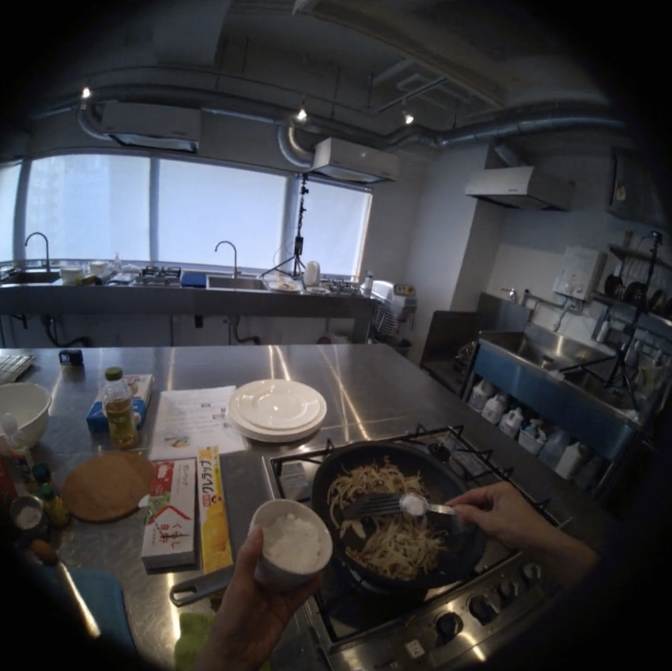}
        \caption{Caption: ``add sugar''}
        \label{fig:fig2}
    \end{subfigure}
    \caption{Example of key steps that have similar video frames but different step ID and language annotations.}
    \label{fig:key_step}
\end{figure}

Table \ref{tab:ego-key-step} summarizes the obtained results, with symbol \(^\dag\) marking that the accuracy is reproduced from \cite{pramanick2023egovlpv2}. The first conclusion is, in terms of acceptance accuracy, \textsf{LATMOS} achieves consistent accurate performance across pre-trained encoders and high-dimensional observation modalities. This is specially relevant because it highlights the modularity \textsf{LATMOS} with respect to modules employed to extract relevant symbols from observations. On the other hand, for the top 2 best vision encoders (Omnivore and EgoVLPv2), the use of the latent representation of \textsf{LATMOS} significantly improves accuracy in the key-step challenge. We claim that the boost is due to the fact that having a meaningful latent representation of the task helps discerning between observations that very close in the observation space but have quite different semantic meaning and step IDs. An instance can be seen in Fig. \ref{fig:key_step}, where two images that are almost equal have different captions.

\begin{table}[ht]
\centering
\caption{Acceptance accuracy and key-step accuracy in the Ego-Exo4D dataset for different pre-trained encoders.}
\label{tab:ego-key-step}
\begin{tabular}{cc|ccc}
\multirow{2}{*}{encoder} &\multirow{2}{*}{modality} & acceptance & \multicolumn{2}{c}{key-step} \\
\cline{4-5}
& & \textsf{LATMOS} & single & \textsf{LATMOS}\\ 
\hline
\multirow{3}{*}{XCLIP \cite{ni2022expandinglanguageimagepretrainedmodels}} & L & 0.901 & 0.865 & 0.852  \\
& V  & 0.681 & 0.353 & 0.387 \\
& VL & 0.902  & 0.806 & 0.626\\
\hline
Omnivore \cite{girdhar2022omnivore} &
V & 0.907 & 0.352 & 0.464\\
\hline
EgoVLPv2\cite{pramanick2023egovlpv2} & 
V & 0.866 & 0.368\(^\dag\) & 0.448
\end{tabular}
\end{table}

\subsection{Robot Task Planning from Learned Automaton}
\label{subsec:doorkey}

\begin{figure}[t]
    \centering
    \includegraphics[width=\linewidth]{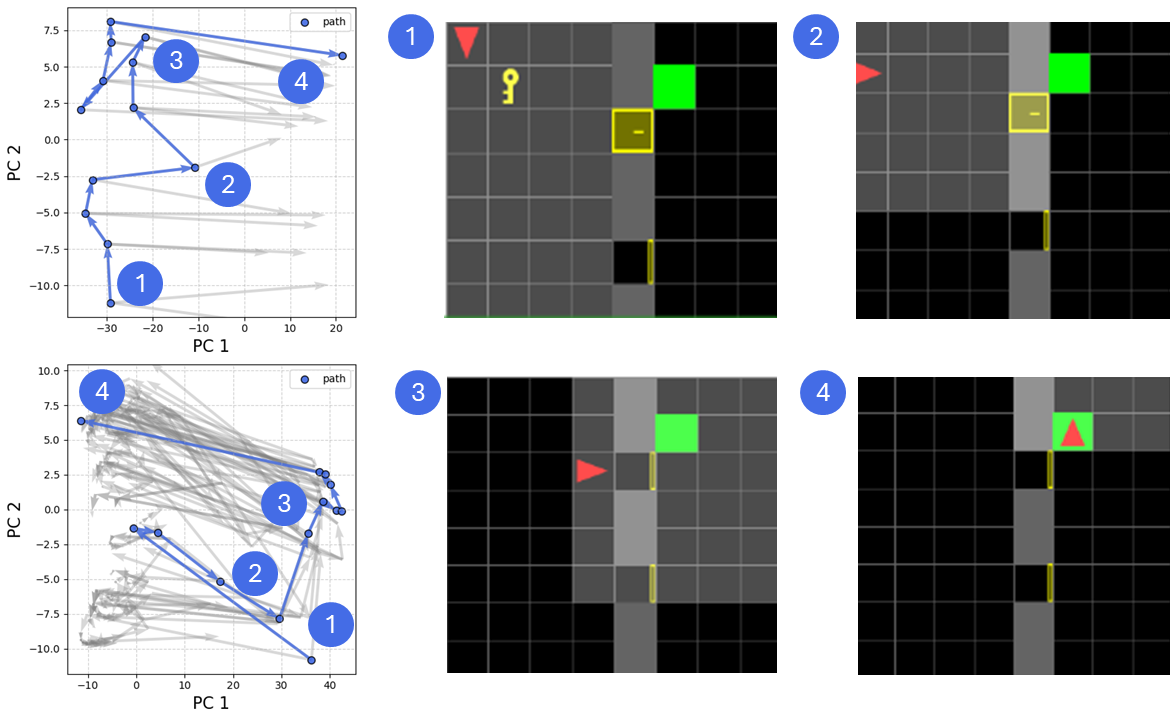}
    \caption{Visualization of the task plans computed by \textsf{LATMOS-v} and \textsf{Dijkstra}. The panels on the left report PCA-projected \textit{latent} state representations of the search tree produced by both heuristics (\textsf{LATMOS-v} on the top, \textsf{Dijkstra} on the bottom), in which, the blue trajectories are the final plan and the gray traces are A* search branches. The panels on the right display Door-Key configurations corresponding to four key states on the final plan (robot in red, goal in green, door in yellow, walls in gray). Both heuristics produced \textit{the same optimal plans from same starting states}, but \mbox{\textsf{LATMOS-v}} is significantly better in search efficiency. }
    \label{fig:doorkey_path}
\end{figure}

After evaluating the model checking capabilities of \mbox{\textsf{LATMOS}}, we assess its task planning capabilities in a 2D grid environment called Door-Key \cite{chevalier2023minigrid}. In this task, a robot must find a key to open a door and reach a goal while avoiding obstacles. We generate training data from 36 procedurally generated \(8\times8\) grid environments with varying key, door, and goal positions. We consider two variants of \textsf{LATMOS}: \textsf{LATMOS-x}, which takes in explicit agent location, key location, goal location and door status; and \textsf{LATMOS-v}, which takes in bird-eye view RGB images of the environment (right part of Fig. \ref{fig:doorkey_path}). The domain-specific encoder of \textsf{LATMOS-v} is a one-layer five-channel convolutional neural network. No encoder is used for \textsf{LATMOS-x}, we directly feed the observation to the sequence model.

For task planning, we compare four heuristics for the model-guided A* planner: a \textsf{Dijkstra} variant corresponding to a zero heuristic; the \textsf{LATMOS-x} heuristic as proposed in \eqref{eq:heuristic}; the \textsf{LATMOS-v} heuristic as proposed in \eqref{eq:heuristic}; and a \textsf{o-l2} heuristic, which is the \(l_2\) heuristic applied to the bird-eye images of the environment. 

\begin{figure}[t]
    \centering
    \begin{tabular}{c}
    \includegraphics[width=0.95\linewidth]{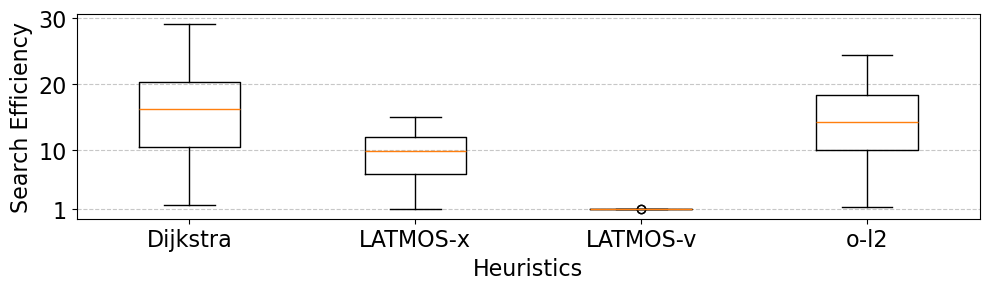}
    \end{tabular}
    \caption{Search efficiency for different heuristics in the Door-Key task. \textsf{LATMOS} significantly outperforms the other heuristics, almost always achieving a successful task plan by building a single-branch tree, i.e, \(E \simeq K^{\prime\prime}\). }
    \label{fig:search_effiency}
\end{figure}

As shown in Fig. \ref{fig:search_effiency}, the two \textsf{LATMOS} heuristics significantly outperform the baseline heuristics. In particular, \textsf{LATMOS-v} often computes a successful task plan by building a single-branch tree, as the ratio between number of explored states and the length of the shortest computed path is near $1$. The better performance of the \textsf{LATMOS} heuristics underlines the fact that our proposed method is able to learn meaningful latent task representations not only for model checking, but also for enhanced planning. The improvement of \textsf{LATMOS-v} over \textsf{LATMOS-x} is because the domain-specific encoder learns embeddings that help with the planning, in contrast to directly learning the arithmetic relationships between robot location, key location, goal location and door status as in \textsf{LATMOS-x}. To better understand the planning features of \textsf{LATMOS}, we provide a qualitative visualization of task plans generated by the \textsf{Dijkstra} and \textsf{LATMOS-v} heuristics in Fig. \ref{fig:doorkey_path}. Although A* generates a valid task plan with both heuristics, \textsf{LATMOS-v} is significantly more efficient than \textsf{Dijkstra}, in accordance to the search efficiencies reported in Fig. \ref{fig:search_effiency}.

%% file: Conclusion.tex
\section{CONCLUSION}\label{sec:conclusion}

This paper proposed \textsf{LATMOS}, an automata-theory-inspired approach for learning task models from positive task demonstrations, which can subsequently be used for robot task planning. By integrating a neural 
network observation encoder with a latent-space sequence model, \textsf{LATMOS} effectively factorizes a task, interprets its execution state, and enables task planning in a continuous latent space. Our experiments show \textsf{LATMOS}'s ability to generalize across various domains, from LTL-specified tasks to real-world video-based task demonstration to robot task planning. Our results indicate that \textsf{LATMOS} is capable of both traditional model checking and efficient task planning using the sequence model for heuristic guidance. Future work will focus on enhancing the interpretability of the latent feature space. This includes exploring feature space alignment techniques with language features, which could enable interpretation and task guidance from large language models (LLMs). Additionally, while our planning method relies on A* in the task space, we recognize the potential for integrating energy-based models \cite{Lecun2024EBM} and optimization-based planning to avoid negative sampling and provide continuous-action planning. Finally, we plan to test \textsf{LATMOS} in real-world robot experiments.